\documentclass{article} % For LaTeX2e
\usepackage{nips15submit_e,times}
\usepackage{hyperref}
\usepackage{url}
\usepackage{cite}

\usepackage{graphicx}
\usepackage{subcaption}
\usepackage{epsfig}
\usepackage{array}
\usepackage{booktabs}
\usepackage{color}
\usepackage{multirow}

\usepackage[cmex10]{amsmath}
\usepackage{amssymb}

\title{Discovery Radiomics via StochasticNet Sequencers for Cancer Detection}

\author{
Mohammad Javad Shafiee$^{*}$, Audrey G. Chung, Devinder Kumar, Alexander Wong \\
Department of Systems Design Engineering, University of Waterloo, ON N2L 3G1 \\
\texttt{$^{*}$ mjshafiee@uwaterloo.ca} \\
\And
Farzad Khalvati, Masoom A. Haider \\
Department of Medical Imaging, Sunnybrook Research Institute, University of Toronto, ON \\
}
\nipsfinalcopy % Uncomment for camera-ready version

\begin{document}

\maketitle

\begin{abstract}
Radiomics has proven to be a powerful prognostic tool for cancer detection, and has previously been applied in lung, breast, prostate, and head-and-neck cancer studies with great success. However, these radiomics-driven methods rely on pre-defined, hand-crafted radiomic feature sets that can limit their ability to characterize unique cancer traits. In this study, we introduce a novel \textit{discovery radiomics} framework where we directly discover custom radiomic features from the wealth of available medical imaging data. In particular, we leverage novel StochasticNet radiomic sequencers for extracting custom radiomic features tailored for characterizing unique cancer tissue phenotype.  Using StochasticNet radiomic sequencers discovered using a wealth of lung CT data, we perform binary classification on 42,340 lung lesions obtained from the CT scans of 93 patients in the LIDC-IDRI dataset.  Preliminary results show significant improvement over previous state-of-the-art methods, indicating the potential of the proposed discovery radiomics framework for improving cancer screening and diagnosis.
\end{abstract}

\vspace{-0.55 cm}
\section{Introduction}
\label{Introduction}
\vspace{-0.35 cm}
Radiomics is a promising and powerful prognostic method for the detection of cancerous tissue. Referring to the high-throughput extraction and analysis of large amounts of quantitative imaging-based features from standardized medical imaging data, radiomics allows for quantitative tumour phenotype characterizations and cancer detection and prognosis via a high-dimensional mineable feature space \cite{Lambin2012}. Radiomics has previously been applied to lung, breast, prostate, and head-and-neck cancer patient cases \cite{Aerts2014} \cite{Gevaert2012} \cite{Khalvati2015} \cite{Maforo2015}, and demonstrated the prognostic power of radiomics and the potential of radiomic features for personalized medicine, risk stratification, and predicting patient outcomes. However, these radiomic-driven methods rely on pre-defined, hand-crafted quantitative features based on intensity, texture, and shape, and as may not be able to fully characterize the unique traits of specific forms of cancer.  As such, a way to uncover quantitative radiomic features tailored for characterizing unique cancer phenotype from standardized imaging data is highly desired.
 In this study, we introduce a novel \textit{discovery radiomics} framework where we bypass conventional predefined, hand-crafted radiomic feature models and directly discover custom radiomic feature models via the abundance of readily available medical imaging data. Discovery radiomics has the potential to find new abstract features that capture unique characteristics of cancer phenotypes beyond what predefined feature models can extract, allowing for improved personalized medicine.
\vspace{-0.4 cm}

\section{Methods}
\label{Methods}
\begin{figure*}[t]
\vspace{- 1 cm}
	\centering
	\includegraphics[width = 0.7\linewidth]{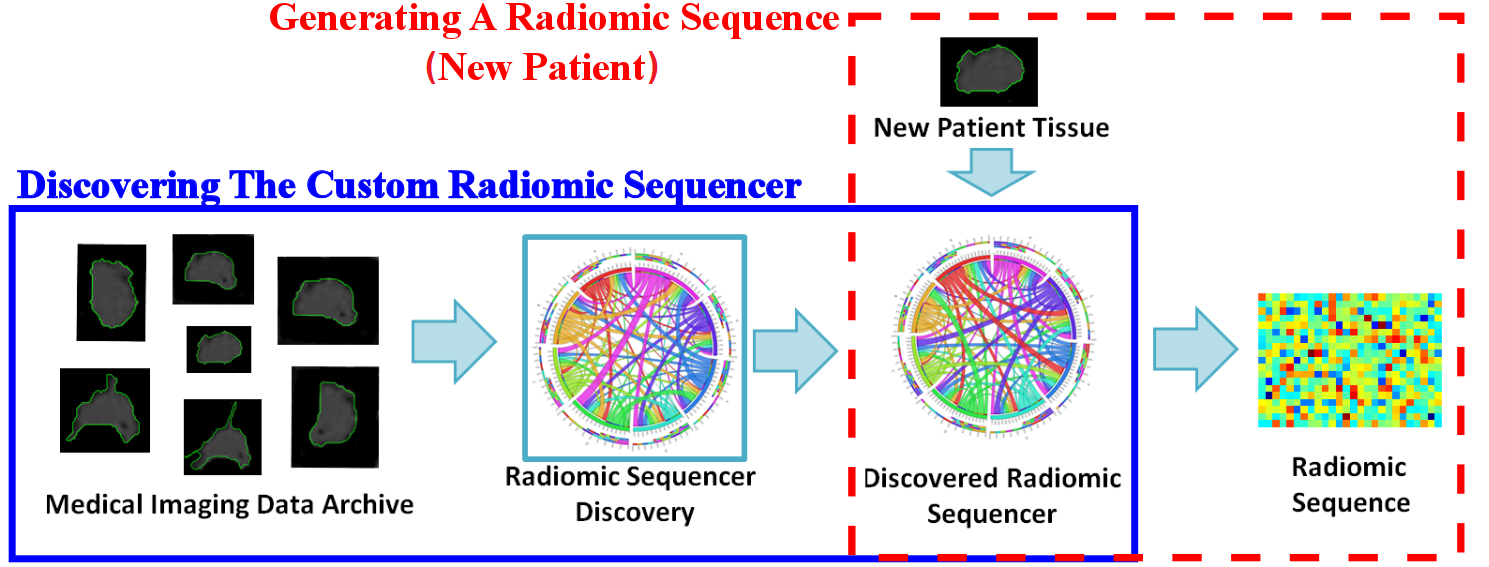}
	\caption{Overview of the proposed discovery radiomics framework for cancer detection. A custom radiomic sequencer is discovered via past medical imaging data; for new patients, radiomic sequences of custom radiomic features are generated for tissue quantification and analysis.}
	\vspace{-10pt}
	\label{fig_AlgFrame}
\end{figure*}

\vspace{-0.4 cm}
The proposed discovery radiomics framework can be described as follows (see Figure~\ref{fig_AlgFrame}). Given past radiology data and corresponding pathology-verified radiologist tissue annotations from a medical imaging data archive, the radiomic sequencer discovery process learns a radiomics sequencer that can extract highly customized radiomic features that are tailored for characterizing unique tissue phenotype that differentiate cancerous tissue from healthy tissue.  The discovered radiomic sequencer can be applied to a new patient data to extract the corresponding radiomic sequence for cancer screening and diagnosis purposes.

The radiomic sequencer being discovered in this study is built upon a deep convolutional StochasticNet~\cite{stochasticnet} architecture, where a deep convolutional neural network (CNN) is represented as a random graph and the neural connections within this network are formed stochastically based on a probabilistic neural connectivity model, thus leveraging random graph theory~\cite{Gilbert} to construct more efficient deep neural network architectures that retain modeling capabilities of traditional, densely-connected network architectures. The radiomic sequencer discovered in this study consists of three stochastically-formed convolutional layers, each containing 32, 32, and 64 receptive fields (size $5 \times 5$), respectively. Each receptive field is part of a realization of a random graph with a uniform neural connectivity probability of 0.5; that is, the expected number of parameters in each receptive field of the proposed sequencer is only half that of a receptive field in a sequencer built using a conventional deep CNN. Less number of parameters and, therefore, more efficient training and faster testing running time are the most important advantages of the proposed framework compared to the conventional CNN approaches.

\vspace{-0.35 cm}

\section{Results and Discussion}
\label{ResultsDiscussion}
\vspace{-0.35 cm}
In this study, we used a subset of the LIDC-IDRI~\cite{LungData1,lungData2} dataset. The CT images were acquired via a broad range of CT scanner models from different manufacturers using the following tube peak potential energies for acquiring the scans: 120$kV$ ($n = 818$), 130$kV$ ($n=31$), 135$kV$ ($n=69$), and 140$kV$ ($n=100$). A subset of 93 patient cases which have definitive diagnostic results was selected and, using data augmentation, an enriched dataset of 42,340 lung lesions was obtained via the rotation of each malignant and benign lesion by 45$^\circ$ and 10$^\circ$ increments, respectively. The proposed framework was evaluated using the enriched dataset and quantitatively compared to two state-of-art methods~\cite{Zinovev2011}~\cite{Kumar2015}.
Note that while a Multi-scale Convolutional Neural Networks (MCNN) architecture~\cite{Shen2015} was recently proposed and achieved an accuracy as high as $86.84\%$ via parameter tuning, this method was trained and tested on radiologist interpretations only and not definite diagnostic results, making the ground truth subject to high inter-observer variability.

As seen in Table~\ref{tab_Improvement} the StochasticNet radiomic sequencer (SNRS) achieves high sensitivity while maintains the good specificity. The reported overall accuracy for SNRS noticeably outperforms the tested state-of-the-art methods.  These preliminary results illustrate the potential of the proposed discovery radiomics framework for improving cancer screening and diagnosis.

\vspace{-0.5 cm}
\begin{table}[!htp]
	\centering
	\caption{Comparison of performance metrics for belief decision trees (BDT)\cite{Zinovev2011}, a deep autoencoding radiomic sequencer (DARS)~\cite{Kumar2015}, and the discovered StochasticNet radiomic sequencer (SNRS).}
	\begin{tabular}{|c|c|c|c|}
	%\begin{tabular}{|p{1.7cm}|p{4.5cm}|p{3cm}|p{3.5cm}|p{1.7cm}|}
		\hline
		& \textbf{Sensitivity} & 	\textbf{Specificity} &		\textbf{Accuracy} \\
		\hline \hline
		\textbf{BDT~\cite{Zinovev2011}}	& 	$N/A$  			&	$N/A$ 			&	$54.32\%$		\\ \hline
		\textbf{DARS~\cite{Kumar2015}}			& 	$83.35\%$		&	20.18 			&	$75.01\%$		\\ \hline		% Specificity = $20.18\%$
		\textbf{SNRS}					& 	$\textbf{91.07\%}$  		&	$\textbf{75.98\%}$ 		&	$\textbf{84.49\%}$		\\ \hline
	\end{tabular}
	\label{tab_Improvement}
\end{table}	
\vspace{-1 cm}
\subsubsection*{Acknowledgments}
\vspace{-0.35 cm}
This research has been supported by the Ontario Institute of Cancer Research (OICR), Canada Research Chairs programs, Natural Sciences and Engineering Research Council of Canada (NSERC), and the Ministry of Research and Innovation of Ontario.  The authors also thank Nvidia for the GPU hardware used in this study through the Nvidia Hardware Grant Program.

\renewcommand{\refname}{\normalfont\selectfont\normalsize\bf References}
\small{
	\bibliographystyle{IEEEtran}
	\bibliography{DiscoveryRadiomics}
}

\end{document}